\documentclass[11pt,a4paper]{article}
\usepackage[hyperref]{emnlp-ijcnlp-2019}
\usepackage{times}
\usepackage{latexsym}

\usepackage{url}
\usepackage{booktabs} 

\usepackage{amsmath}
\usepackage{amsfonts}
\usepackage{bm}
\usepackage{array}
\usepackage{enumitem}
\usepackage{needspace}
\usepackage{adjustbox}
\usepackage{tikz}
\usepackage{pgfplots}
\usepackage{multirow}
\usepackage{autobreak}
\usepackage{makecell}
\usepackage{xcolor}
\usepackage{graphicx}
\usepackage{balance}
\usepackage{amssymb}
\usepackage{import}
\usepackage{caption}
\usepackage{subcaption}
\usepackage[symbol]{footmisc}
\usepackage{amssymb}
\usepackage{nicefrac}       
\usepackage{microtype}      

\aclfinalcopy 

\newcommand{\changed}[1]{{#1}}

\title{Adaptive Parameterization for Neural Dialogue Generation}

\author{
Hengyi Cai$^{\dagger,\S}$\thanks{\ \ Work done at Data Science Lab, JD.com.}, Hongshen Chen$^\ddagger$, Cheng Zhang$^\dagger$, Yonghao Song$^\dagger$, Xiaofang Zhao$^\dagger$ {\normalfont{and}} Dawei Yin$^\ddagger$ \\
$^\dagger${Institute of Computing Technology, Chinese Academy of Sciences} \\
$^\S${University of Chinese Academy of Sciences} \\
$^\ddagger${Data Science Lab, JD.com} \\
{caihengyi@ict.ac.cn, ac@chenhongshen.com,} \\
{\{zhangcheng, songyonghao, zhaoxf\}@ict.ac.cn, yindawei@acm.org}
}

\date{}

\begin{document}
\maketitle
\begin{abstract}

 Neural conversation systems generate responses based on the sequence-to-sequence (SEQ2SEQ) paradigm.
 Typically, the model is equipped with a single set of learned parameters to generate responses for given input contexts. 
 When confronting diverse conversations, its adaptability is rather limited and the model is hence prone to generate generic responses.
 In this work, we propose an {\bf Ada}ptive {\bf N}eural {\bf D}ialogue generation model, \textsc{AdaND}, which manages various conversations with conversation-specific parameterization.
 For each conversation, the model generates parameters of the encoder-decoder by referring to the input context.
 In particular, we propose two adaptive parameterization mechanisms: a context-aware and a topic-aware parameterization mechanism.
 The context-aware parameterization directly generates the parameters by capturing local semantics of the given context.
 The topic-aware parameterization enables parameter sharing among conversations with similar topics by first inferring the latent topics of the given context and then generating the parameters with respect to the distributional topics.
 Extensive experiments conducted on a large-scale real-world conversational dataset show that our model achieves superior performance in terms of both quantitative metrics and human evaluations.

\end{abstract}
\section{Introduction}

 Open-domain dialogue models, usually called chit-chat systems, draw increasing attention from both academia and industry in recent years.
 Building on the successful sequence-to-sequence (SEQ2SEQ) paradigm~\citep{DBLP:journals/corr/SutskeverVL14,DBLP:journals/corr/ChoMGBSB14,Bahdanau2014NeuralMT},
 contemporary mainstream open-domain dialogue generation models~\cite{DBLP:conf/acl/ShangLL15, DBLP:conf/aaai/SerbanSBCP16, DBLP:journals/corr/SerbanSLCPCB16, DBLP:conf/acl/ShenSLLNZAL17, DBLP:conf/eacl/ClarkC17, DBLP:conf/acl/ZhaoZE17,  DBLP:conf/aaai/XingWWLHZM17, LAED}, trained on a large number of context-response pairs, attempt to generate an appropriate response for the given context based on a single set of the model parameters.

 Because of its great potential in understanding and modeling conversations, SEQ2SEQ has been widely applied in different kinds of conversation scenarios including technical supporting, movie discussions, and social entertainment, etc. 
 However, when confronting conversations with diverse topics or themes, SEQ2SEQ is usually prone to make generic meaningless responses due to its oversimplified parameterization.
 To tackle this issue, \citet{DBLP:conf/aaai/XingWWLHZM17} proposed a topic-aware response generation model, which utilizes a single encoder-decoder, augmented with topic information obtained from a pre-trained LDA model.
 Though effective, the model heavily relies on the outsourcing topic information to capture the topic variations of different conversations.
 Another approach, per-topic/theme encoder-decoder model~\citep{DBLP:journals/corr/abs-1708-00897}, uses separate encoder-decoder model for each topic or theme.
 This method needs the preorganized topic/theme annotations for each conversation, which are prohibitively expensive to obtain.
 Furthermore, building multiple separate topic/theme-specific encoder-decoders not only weakens the applicability and efficiency of the system, but also prevents parameter sharing across domains, which leads to overparameterization due to the excessive amount of model parameters.

 To gather the benefits of both approaches, in this paper, we propose an adaptive neural dialogue generation model which utilizes a single encoder-decoder for diverse conversations, meanwhile, the encoder-decoder is specifically parameterized according to each conversation.
 In particular, we propose two adaptive parameterization mechanisms: 
 1) context-aware parameterization directly generates parameters of the encoder-decoder model by capturing local semantics of the given context;
 2) topic-aware parameterization enables parameter sharing among conversations with similar topic distributions by first inferring the latent topics of the input context, and then generating the parameters with respect to the inferred latent topics.
 Equipped with both the context-aware and topic-aware parameterization mechanisms, the model is capable of generating responses for diverse conversations with a single encoder-decoder through a more flexible and efficient approach.
 Moreover, our model is trained in an end-to-end fashion without any costly external or labeled topic annotations.
 
 We empirically evaluate our approach on a large scale real-world conversational dataset.
 Extensive experiments show that our proposed \textsc{AdaND} outperforms existing dialogue generation models in terms of both the automatic evaluation metrics and human judgements.

\section{Neural Dialogue Generation Model}

Conventionally, neural dialogue generation model follows the encoder-decoder paradigm.
Given the context $\mathbf{x}=\{x_1, x_2, \cdots, x_{T_\mathbf{x}}\}$ and the target response $\mathbf{y}=\{y_1, y_2, \cdots, y_{T_{\mathbf{y}}}\}$, the model learns to maximize the following conditional probability:
\begin{align}
    p(\mathbf{y}|\mathbf{x}) = \prod_{t=1}^{T_{\mathbf{y}}}p(y_t|\mathbf{y}_{<t}, \mathbf{x}),
\end{align}
where $\mathbf{y}_{<t}=y_1\cdots{}y_{t-1}$. 

Typically, the probability $p(y_t|\mathbf{y}_{<t}, \mathbf{x})$ is modeled as follows:
\begin{align}
    p(y_t|\mathbf{y}_{<t}, \mathbf{x}) = p(y_t|s_{t-1}),
\end{align}
where $s_{t-1}$ is the decoding hidden state up to time step $t-1$, depending on $\mathbf{y}_{<t}$ and $\mathbf{x}$.
The $s_t$ can be defined as:
\begin{align}
    s_t &= \mathcal{F}(y_t, s_{t-1}, \mathcal{G}(\mathbf{x})),
\end{align}
where the encoder $\mathcal{G}$ and the decoder $\mathcal{F}$ can be implemented as recurrent neural networks such as LSTMs~\cite{Hochreiter:1997:LSTM:1246443.1246450} or GRUs~\cite{DBLP:journals/corr/ChoMGBSB14},
or the transformer~\cite{transformer} (with the attention mechanism~\cite{Bahdanau2014NeuralMT} or self-attention mechanism).

In this work, we employ the LSTM-based encoder-decoder dialogue generation model.
The LSTM unit is formulated as:
\begin{align}
\begin{split}
    i_t &= W_h^i{}h_{t-1} + W_x^i{}x_t + b^i \\
    g_t &= W_h^g{}h_{t-1} + W_x^g{}x_t + b^g \\
    f_t &= W_h^f{}h_{t-1} + W_x^f{}x_t + b^f \\
    o_t &= W_h^o{}h_{t-1} + W_x^o{}x_t + b^o \\
    c_t &= \sigma(f_t) \odot c_{t-1} + \sigma(i_t) \odot \text{tanh}(g_t) \\
    h_t &= \sigma(o_t) \odot \text{tanh}(c_t),
\end{split}
\end{align}
where $\sigma$ is the sigmoid operator and $\odot$ stands for Hadamard product.
$h_{t-1}$ is the previous hidden state and $x_t$ is the input embedding at step $t$.
$W=\{W_h^i, W_h^g, W_h^f, W_h^o, W_x^i, W_x^g, W_x^f, W_x^o\}$ and $b = \{b^i, b^g, b^f, b^o\}$ are the LSTM parameters.
 The model parameters are tuned on the training corpus.
 
 When testing, given the input context, it generates a response with the learned parameters.
 This architecture shows great success in neural dialogue generation~\cite{DBLP:conf/acl/ShangLL15, DBLP:conf/aaai/SerbanSBCP16, DBLP:journals/corr/SerbanSLCPCB16, DBLP:conf/acl/ShenSLLNZAL17, DBLP:conf/eacl/ClarkC17, DBLP:conf/aaai/XingWWLHZM17, DBLP:journals/corr/abs-1708-00897, LAED}.
 However, with a single set of model parameters and the oversimplified model architecture, the flexibility of the model is rather limited, especially when confronting conversations with diverse topics or themes.

\begin{figure}[t!]
  \centering
  \includegraphics[width=0.46\textwidth]{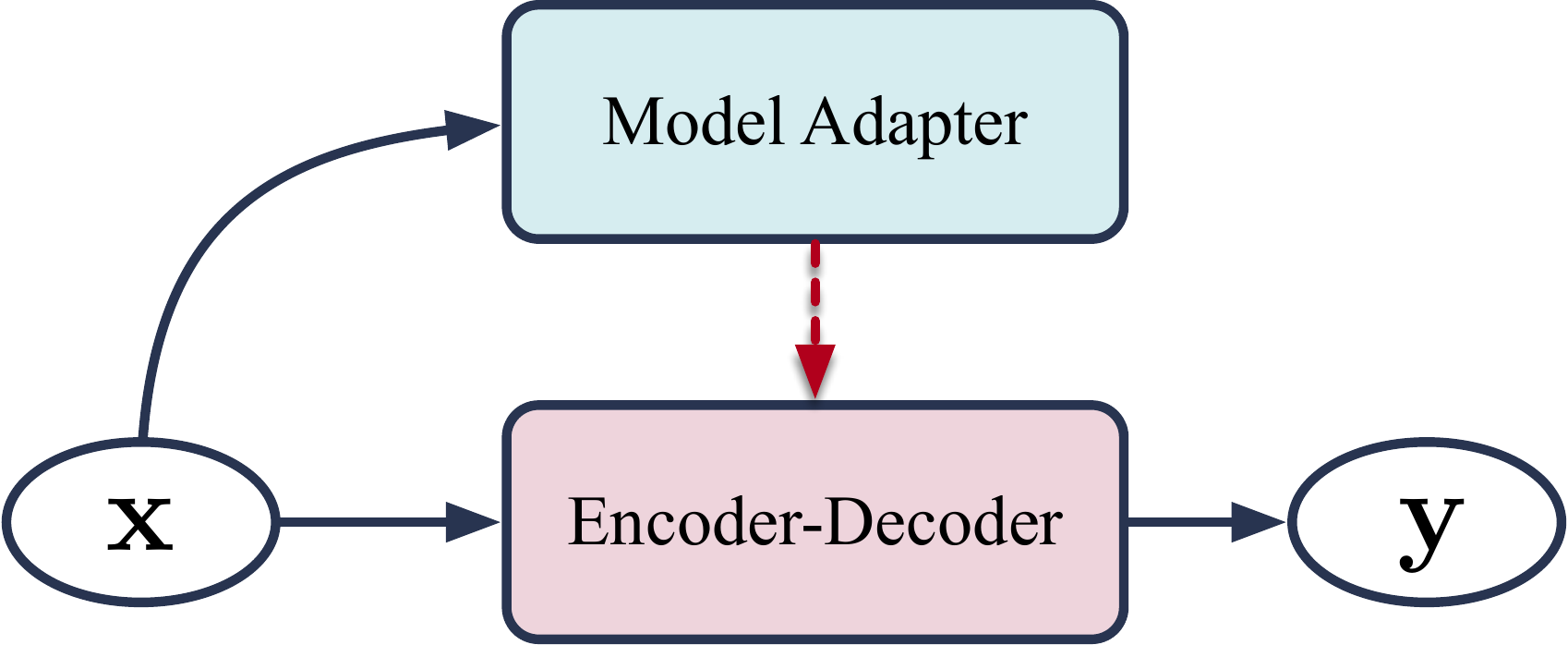}
  \caption{
    General model architecture.
    Black solid lines denote information flow, and the red dashed line indicates the adaptive parameterization operation.
  }
  \label{fig:model_arch}
\end{figure}

\section{Adaptive Neural Dialogue Generation Model}
 In this work, we propose an adaptive neural dialogue generation model which utilizes a single encoder-decoder model and a set of dynamical parameters to balance the model's flexibility and efficiency.
As depicted in Figure~\ref{fig:model_arch}, we utilize the model adapter to parameterize the encoder-decoder for each conversation. 
It takes the dialogue context as its input, and generates parameters of the encoder-decoder model through two adaptive parameterization mechanisms;
and then the resultant encoder-decoder model generates the response with a specific set of model parameters.

\begin{figure}[!t]
  \centering
  \includegraphics[width=0.48\textwidth]{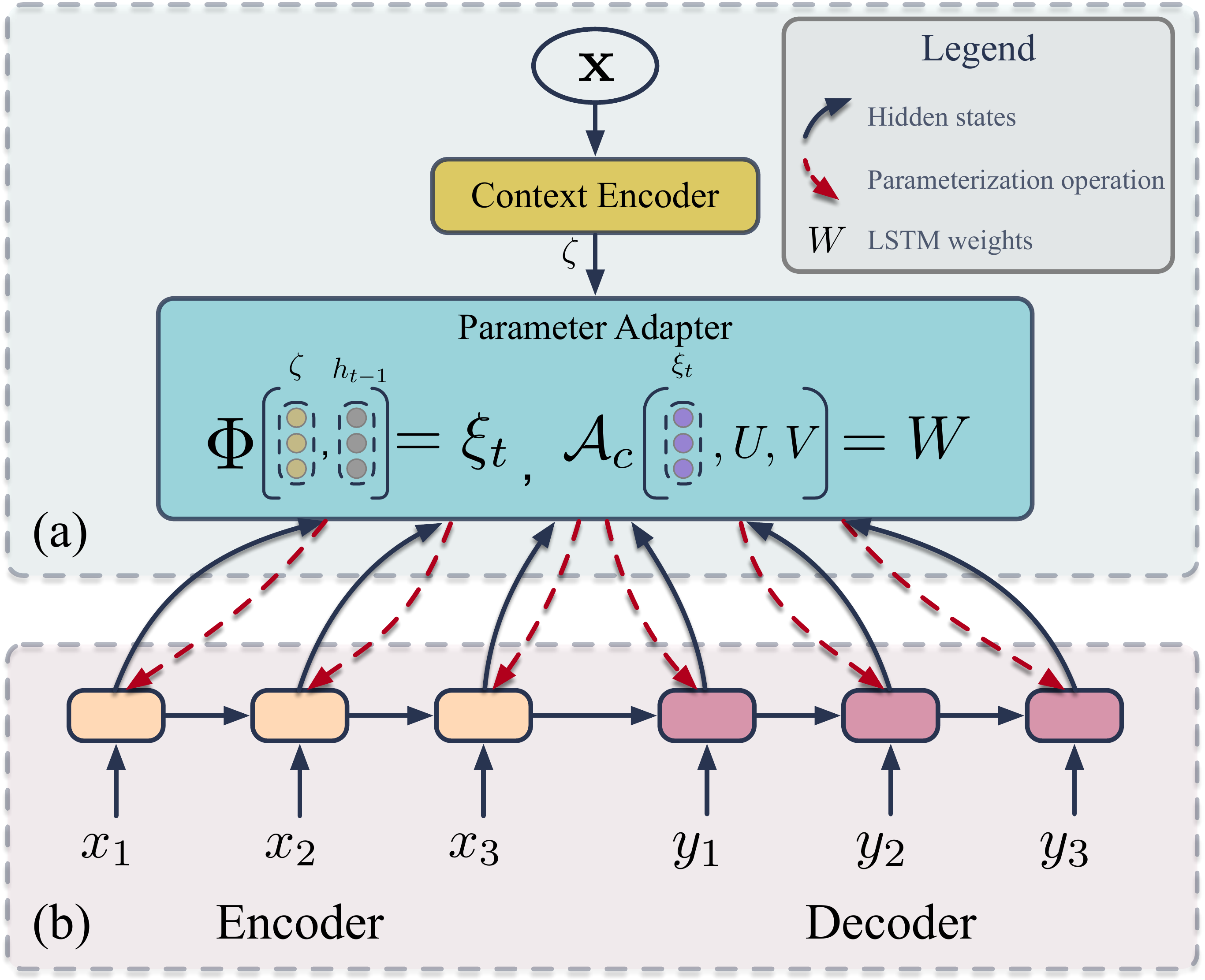}
  \caption{
    General architecture of the context-aware parameterization.
    (a) Context-aware parameter adapter. (b) One layer of the dialogue generation model.
    Black solid lines denote information flow, and red dashed lines indicate adaptive parameterization operations.
    $\mathcal{A}_c$ stands for the context-aware parameterization function (Eq.(\ref{eq:context_svd})).
  }
  \label{fig:CDA}
\end{figure}

\subsection{Context-aware Parameterization}

 Context-aware parameterization parameterizes the dialogue encoder-decoder with respect to the local semantics of the given context.
 As shown in Figure~\ref{fig:CDA}, we first obtain the semantic representation of the context through a context encoder. 
 Then, the parameter adapter dynamically adapts the parameters of the encoder-decoder at each time step.
 Here, we utilize a bidirectional LSTM to transform the input context into the semantic hidden representation $\zeta$. 

 The parameter adapter then generates the weights of LSTM units as:
 \begin{equation}
        W = M^{[1:N_r]}\zeta{},
 \end{equation}
 where $W\in\mathbb{R}^{N_r\times{}N_c}$ and $M^{[1:N_r]}\in\mathbb{R}^{N_r\times{}N_c\times{}N_\zeta}$ is a tensor.
 $N_r=4N_h$ and $N_c=N_h+N_x$, in which $N_h$ is the hidden size of the LSTM and $N_x$ is the embedding size.
 $\zeta\in\mathbb{R}^{N_\zeta}$ is the context representation.
 The product $M^{[1:N_r]}\zeta$ results in a weight matrix $W$, where each row is computed by one slice $j=1,2,\cdots,N_r$ of the tensor: $W_j=(M^{j}\zeta)^{\mathrm{T}}$.

 Although such parameterization seems to be straightforward, due to the quadratic size of $M^{[1:N_r]}$, the parameter size of such parameter adapter is $N_\zeta$ times larger than the basic encoder-decoder model. Therefore, overfitting and expensive computational cost make it infeasible~\cite{DBLP:conf/nips/BertinettoHVTV16, DBLP:journals/corr/HaDL16}.
 Following \citet{DBLP:conf/nips/FlennerhagYKE18}, we reduce the parameter space by factorizing the weights as:
 \begin{align}
    \begin{split}
        W &= \mathcal{A}_c(\xi_t, U, V) \\
        \xi_t &= \Phi(\zeta, h_{t-1}), \\
    \end{split}
 \end{align}
 where $\Phi$ is implemented as a LSTM unit, 
 $h_{t-1}$ is the previous encoder or decoder hidden state, and $\xi_t\in\mathbb{R}^{N_\zeta}$ is the context representation at time step $t$.
 $\mathcal{A}_c$ denotes the context-aware parameterization function, defined as:
 \begin{equation}
     \mathcal{A}_c(\xi_t, U, V) = U\xi_t{}V^{\mathrm{T}},
     \label{eq:context_svd}
 \end{equation}
 where $U\in{\mathbb{R}}^{N_r\times{N_\zeta}}$ and $V\in\mathbb{R}^{N_c\times{N_\zeta}}$ are learnable weights.

 The context-aware parameterization function $\mathcal{A}_c$ is reminiscent of the Singular Value Decomposition (SVD).
 Here, $\mathcal{A}_c$ composes a projection by adapting the dialogue context to the weight matrices and does not perform matrix decomposition actually.
 The number of parameters in this formulation is $L=N_r\times{}N_\zeta+N_c\times{}N_\zeta$ and the total parameter number of the model is linear with $L$, so that the total number of model parameters does not explode.

\subsection{Topic-aware Parameterization}
\begin{figure}[t!]
  \centering
  \includegraphics[width=0.48\textwidth]{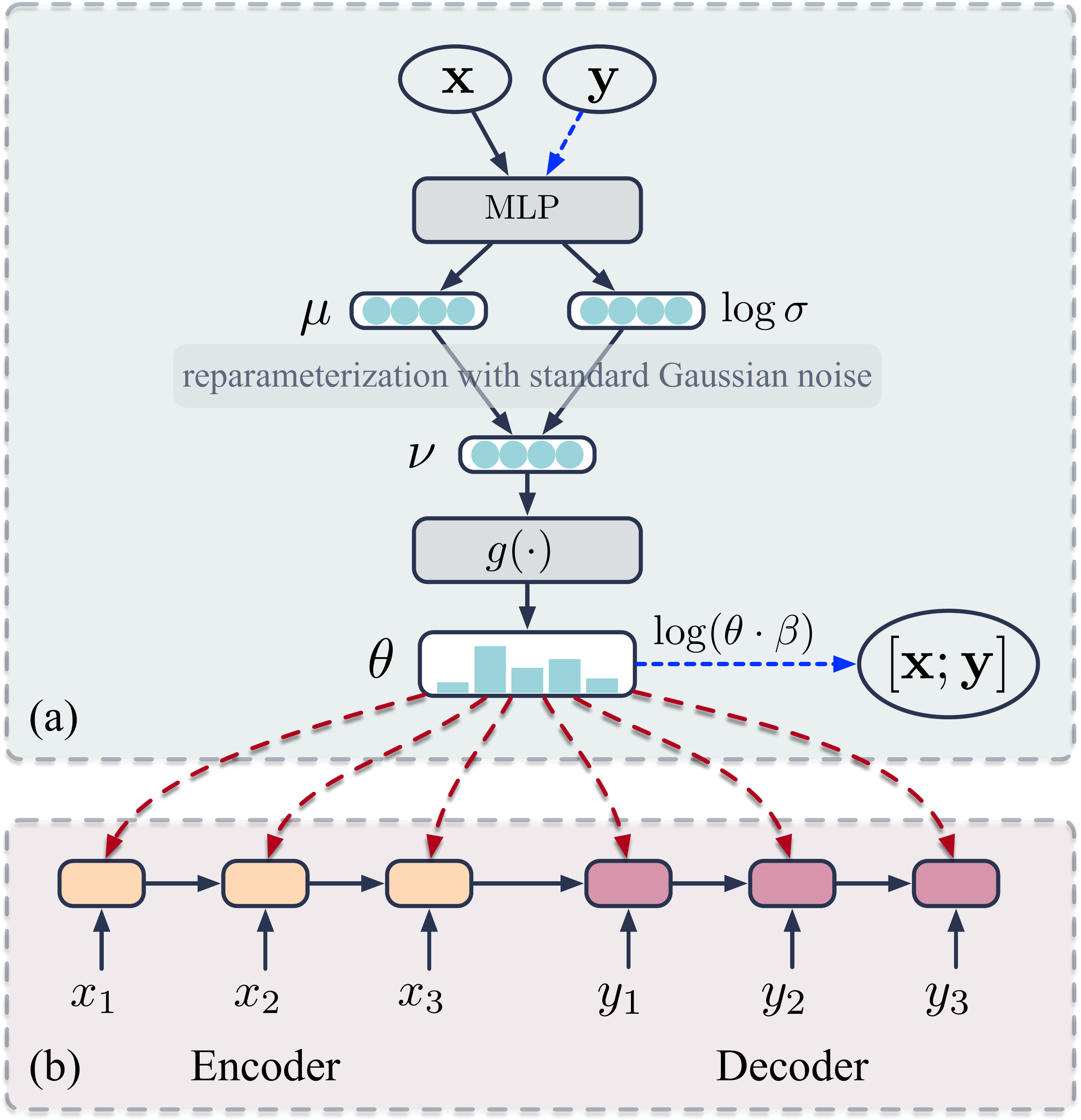}
  \caption{
    Illustration of the topic-aware parameterization.
    (a) Topic inferrer. (b) One layer of the dialogue generation model. 
    Black solid lines denote information flow, blue dashed lines only appear in training, and red dashed lines indicate adaptive parameterization operations.
  }
  \label{fig:TDA}
 \end{figure}

 Context-aware parameterization adapts the encoder-decoder parameters regarding each input context.
 As a result, the adapted encoder-decoder is sensitive to the input context. 
 To enable the parameter sharing among similar topics, we further propose a topic-aware parameterization mechanism.
 As shown in the Figure~\ref{fig:TDA}, the topic inferrer first distills the topic distribution $\theta$ from the context (Figure~\ref{fig:TDA} (a)), and then the parameters of the encoder-decoder model are constructed upon $\theta$ (Figure~\ref{fig:TDA} (b)).
 
\subsubsection{Latent Topic Inference}
 We introduce a variational topic inferrer to infer the topic distribution $\theta$ of the conversation.
 Drawing inspirations from neural topic model~\cite{pmlr-v70-miao17a}, as illustrated in Figure~\ref{fig:TDA} (a), the generation process of the variational topic inferrer is formulated as follows:
 \begin{enumerate}[label=(\roman*)]
	\item A latent variable $\nu$ is inferred to convey the underlying semantics of the given context. 
	\item The topic distribution $\theta$ is constructed from the latent variable $\nu$ through a softmax function.
	\item The dialogue $\mathbf{d}$, composing of a context-response pair $(\mathbf{x},\mathbf{y})$, is drawn from the topic distribution over words $p(w_{i}|\beta_{z_i})$, where $z_i$ is a topic assignment sampled from a multinomial distribution parameterized by the topic distribution $\theta$, and $\beta_{z_i}$ is the topic-word distribution of topic assignment $z_i$.
 \end{enumerate}

 Given a context $\mathbf{x}$, the latent variable $\nu$ is sampled from $P(\nu|\mathbf{x})=\mathcal{N}(\mu_{prior}, \sigma^2_{prior})$, and $\mathcal{N}(\mu_{prior}, \sigma^2_{prior})$ is the multivariate normal distribution with mean $\mu_{prior}$ and diagonal covariance matrix $\sigma^2_{prior}$.
 In practice, $\nu$ is reparameterized as: $\nu = \mu_{prior} + \epsilon \cdot \sigma_{prior}$ and $\epsilon$ is a standard Gaussian noise.
 The Gaussian parameters $\mu_{prior}$ and  $\sigma^2_{prior}$ are the outputs of multilayer perceptrons (MLP) given the bag-of-words (BoW) representation of the context as input.
 \changed{To reduce the encoding noise of stop words, here we choose the BoW instead of LSTM for context representations, following \citet{pmlr-v70-miao17a}.}
 To implement neural variational inference, we utilize an inference network $Q(\nu|\mathbf{d})=\mathcal{N}(\mu_{posterior}, \sigma^2_{posterior})$ to approximate the intractable true posterior $p(\nu|\mathbf{d})$, where $\mu_{posterior}$ and $\sigma^2_{posterior}$ are computed in the same way as the prior, taking the bag-of-words representation of dialogue $\mathbf{d}$ as input. 
 The dialogue $\mathbf{d}$ consists of the context-response pair $(\mathbf{x},\mathbf{y})$. 

 The topic distribution $\theta$ is constructed from the latent variable $\nu$ through a softmax function:
 \begin{equation}
    \begin{aligned}
        \theta = g(\nu) = \text{softmax}(\nu\cdot{\mathbf{W}_\nu}),
    \end{aligned}
 \end{equation}
 where $\mathbf{W}_\nu$ is a linear transformation and the bias terms are left out for brevity.

 Then, the dialogue $\mathbf{d}$ is generated based on the topic distribution $\theta$.
 Given $\theta$, the marginal likelihood of a dialogue $\mathbf{d}$ is formulated as:
 \begin{equation}
    p(\mathbf{d}) = \int_\theta{}p(\theta)\prod_{i=1}^{|\mathbf{d}|}\sum_{z_i}p(w_i|\beta_{z_i})p(z_i|\theta)d\theta.
    \label{eq:p_d}
\end{equation}
 In addition, the topic assignment $z_i$ can be integrated out and the log-likelihood of a word $w_{i}$ in dialogue $\mathbf{d}$ can be factorized as:
 \begin{equation}
 \begin{aligned}
 \log p(w_{i}|\beta, \theta)&=\log\sum_{z_i}[p(w_i|\beta_{z_i})p(z_i|\theta)] \\
 &= \log(\theta\cdot\beta^{\mathrm{T}}).
 \end{aligned}
 \label{eq:beta_theta}
 \end{equation}
 The topic-word distribution $\beta_k$ is defined by:
 \begin{equation}
 \beta_k = \text{softmax}(\digamma\cdot{\Lambda_{k}^{\mathrm{T}}}),
 \label{eq:beta_k}
 \end{equation}
 where ${\digamma} \in \mathbb{R}^{C\times{H}}$ is the topical word embedding matrix and $\Lambda \in \mathbb{R}^{K\times{H}}$ is the topic embedding matrix, $K$ is the number of topics, $C$ is the number of topical words, $H$ is the embedding size and $\beta=\{\beta_{1}, \beta_{2},\cdots,\beta_{K} \}$.

\subsubsection{Parameterization with Latent Topics}
 We parameterize the encoder-decoder with the inferred topic distribution $\theta$.
 In context-aware parameterization, the parameters of the encoder-decoder are adapted dynamically at each time step, whereas in topic-aware parameterization, as illustrated in Figure~\ref{fig:TDA} (b), we generate only one set of parameters for each conversation.
 
 Similar to the context-aware parameterization function in Eq.(\ref{eq:context_svd}), given the topic distribution $\theta$, the topic-aware parameterization function $\mathcal{A}_\kappa$ constructs the LSTM weight $W$ as follows:
 \begin{align}
  \begin{split}
    W &= \mathcal{A}_\kappa(\theta, U, V) \\
    \mathcal{A}_\kappa(\theta, U, V) &= U\theta{}V^{\mathrm{T}}, \\
  \end{split}
  \label{eq:topic_svd}
  \end{align}
 where $U\in{\mathbb{R}}^{N_r\times{K}}$ and $V\in\mathbb{R}^{N_c\times{K}}$ are learnable parameters.
 $K$ is the number of latent topics.

\subsection{Parameterization with Both Context and Topics}
 Intuitively, context-aware parameterization is more adept at capturing local semantics of the input context while topic-aware parameterization enables parameter sharing between conversations with similar topic distributions.
 To benefit the model parameterization with both the local and global information, we further adapt  parameters of the encoder-decoder by utilizing both the context representations $\xi_t$ and the topic distribution $\theta$.
 In particular, the LSTM weight $W$ at time step $t$ is adapted as follows:
 \begin{align}
    \begin{split}
        W =& \Psi_t\mathcal{A}_c(\xi_t, U_c, V_c) + \\
           &(1-\Psi_t)\mathcal{A}_\kappa(\theta, U_\kappa, V_\kappa) \\
        \Psi_t =& \sigma(\xi_t, \theta),
    \end{split}
 \end{align}
 where $\Psi_t$ is the gating function deciding whether the parameterization relies more on the context or the topics.
 $U_c, V_c, U_\kappa$ and $V_\kappa$ are learnable weights.
 $\mathcal{A}_c$ and $\mathcal{A}_\kappa$ denote the context-aware and topic-aware parameterization function respectively.
 $\sigma$ is the sigmoid function.

\subsection{Learning}
To enable the joint optimization of latent topic inference, adaptive model parameterization, and response generation in \textsc{AdaND},
given the definitions in Eq.(\ref{eq:p_d}), similar to \citet{DBLP:journals/corr/KingmaW13} and \citet{pmlr-v70-miao17a}, we derive a variational lower bound for the generation likelihood:
\begin{equation}
 \small{
 \begin{aligned}
    \mathcal{J} 
    =& \mathbb{E}_{Q(\theta|\mathbf{d})}[\sum_{i=1}^{|\mathbf{d}|}[\log{\sum_{z_i}[p(w_i|\beta_{z_i})p(z_i|\theta)]}] + \log{p(\mathbf{y}|\mathbf{x})}] \\ 
    &- D_{KL}(Q(\theta|\mathbf{d})||P(\theta|\mathbf{x})) \\
    =& \mathbb{E}_{Q(\theta|\mathbf{d})}[\sum_{i=1}^{|\mathbf{d}|}\log{p(w_i|\beta, \theta)} + \sum_{t=1}^{T_\mathbf{y}}\log{p(y_t|s_{t-1})}] \\
    &- D_{KL}(Q(\theta|\mathbf{d})||P(\theta|\mathbf{x})) \\
    =& \mathbb{E}_{Q(\nu|\mathbf{d})}[\sum_{i=1}^{|\mathbf{d}|}\log{p(w_i|\beta, \nu)} + \sum_{t=1}^{T_\mathbf{y}}\log{p(y_t|s_{t-1})}] \\
    &- D_{KL}(Q(\nu|\mathbf{d})||P(\nu|\mathbf{x})) \\
    \approx & \sum_{i=1}^{|\mathbf{d}|}\log{p(w_i|\beta, \nu)} + \sum_{t=1}^{T_\mathbf{y}}\log{p(y_t|s_{t-1})} \\
    &- D_{KL}(Q(\nu|\mathbf{d})||P(\nu|\mathbf{x})),
 \end{aligned}
 }
 \label{eq:L_d}
\end{equation}
 where $\mathbf{y}=\{y_1, y_2, \cdots, y_{T_y}\}$,
 $P(\nu|\mathbf{x})$ is the prior estimation of the latent variable $\nu$ which approximates the posterior $Q(\nu|\mathbf{d})$.
 The prior $P(\theta|\mathbf{x})$ = $P(g(\nu)|\mathbf{x})$ = $P(\nu|\mathbf{x})$, and the posterior $Q(\theta|\mathbf{d})$ = $Q(g(\nu)|\mathbf{d})$ = $Q(\nu|\mathbf{d})$.
 The first term is the dialogue generation objective in the latent topic inferrer, the second term is the response generation objective, and the third term is the KL divergence between two Gaussian distributions.
 All the parameters are learned by optimizing Eq.(\ref{eq:L_d}) and updated with back-propagation.
 
 The following previously proposed strategies~\cite{DBLP:conf/conll/BowmanVVDJB16,DBLP:conf/acl/ZhaoZE17} are adopted in training to alleviate the vanishing latent variable problem: 
 (1) KL annealing: the weight of the KL divergence term is gradually increasing from 0 to 1 during training;
 (2) Bag-of-words loss: the bag-of-words loss requires the latent variable $\nu$, together with the dialogue context, 
 to reconstruct the response bag-of-words representation $y_{b}$.
\begin{table*}[!ht]
\centering
\scalebox{0.90}{
\begin{tabular}{@{}lcccccccc@{}}
\toprule
\multirow{2}{*}[-0.5em]{\textbf{Models}} & \multicolumn{4}{c}{\textbf{Relevance~(\%)}}    & \multicolumn{3}{c}{\textbf{{Informativeness~(\%)}}}  \\ 
\cmidrule(lr){2-5}
\cmidrule(lr){6-8}
& \textbf{BLEU} & \textbf{Average} & \textbf{Greedy} & \textbf{Extrema} &  \textbf{Distinct-1} & \textbf{Distinct-2} & \textbf{Distinct-3}  \\ \midrule
\textbf{SEQ2SEQ}                   & 0.845  & 69.60 & 64.94  & 45.29  & 0.2822   & 0.5922 & 0.7873 \\ 
\textbf{CVAE}                   & 1.546  & 71.23 & 66.67  & 47.14  & 0.5465   & 1.716  & 2.731  \\
\textbf{LAED}                   & 0.7545 & 69.91 & 63.55  & 43.12  & 0.3890   & 0.9165 & 1.243  \\ 
\textbf{TA-SEQ2SEQ}             & 1.465  & 72.47 & 65.9   & 45.19  & 0.3593   & 0.7994 & 1.016  \\ 
\textbf{DOM-SEQ2SEQ}            & 1.189  & 74.42 & 66.6   & 48.47  & 0.4977   & 1.294  & 1.814  \\ \hline 
\textbf{\textsc{AdaND} (\textit{with context para.})}   & 1.94  & 74.03 & 66.76  & 49.23  & 0.6493 & 1.889 & 2.745  \\ 
\textbf{\textsc{AdaND} (\textit{with topic para.})}     & \textbf{2.051} & 74.17 & 66.65  & 49.04  & 0.5919  & 1.699 & 2.438  \\
\textbf{\textsc{AdaND} (\textit{with both})}       & 1.90 & \textbf{75.59} & \textbf{67.25}  & \textbf{51.17}  & \textbf{0.7092} & \textbf{2.10} & \textbf{3.108}  \\ 
\bottomrule
\end{tabular}
}
\caption{Quantitative evaluation results (\%).}
\label{tbl:main_exp_res}
\end{table*}

\section{Experiments}
\label{sec:exp}


\subsection{Dataset and Competitor Baselines}

To ascertain the effectiveness of the proposed model, we construct an open-domain conversation corpus covering a broad range of resources including a movie discussions dataset collected from Reddit~\citep{DBLP:journals/corr/DodgeGZBCMSW15}, an Ubuntu technical corpus~\citep{DBLP:conf/sigdial/LowePSP15}, and a chit-chat dataset~\citep{DBLP:conf/acl/KielaWZDUS18}.
87,468 context-response pairs were sampled for training, 4,460 for validation and 4,468 for testing.

The code and corpus are available at \url{http://github.com/hengyicai/AdaND}.

The following state-of-the-art models are adopted as our comparison systems.
    \paragraph{SEQ2SEQ}
    The attention-based sequence-to-sequence model~\citep{Bahdanau2014NeuralMT},
    which is a representative baseline.
    \paragraph{CVAE}
    A latent variable conversation model in which it incorporates a latent variable at the sentence-level to inject stochasticity and diversity~\citep{DBLP:conf/eacl/ClarkC17,DBLP:conf/acl/ZhaoZE17}. 
    \paragraph{LAED} A recurrent encoder-decoder conversation model using discrete latent actions for interpretable neural dialogue generation~\citep{LAED}.
    \paragraph{TA-SEQ2SEQ} TA-SEQ2SEQ incorporates the outsourcing topic information into the response generation, where the topics are learned from a separate LDA model to enrich the context~\citep{DBLP:conf/aaai/XingWWLHZM17}.
    \paragraph{DOM-SEQ2SEQ} A domain-aware conversation model consisting of multiple domain-targeted SEQ2SEQ models for response generation~\citep{DBLP:journals/corr/abs-1708-00897}.

\subsection{Evaluation}
Following the evaluation procedure in previous work~\cite{DBLP:conf/naacl/LiGBGD16, DBLP:conf/aaai/XingWWLHZM17, chen2018hierarchical}, experimental results of all models are reported in terms of the relevance and informativeness.
To evaluate the semantic relevance between the generated response and the ground-truth response,  we adopted the BLEU metric~\cite{DBLP:conf/acl/PapineniRWZ02} and three embedding-based similarity metrics proposed in \citet{DBLP:conf/emnlp/LiuLSNCP16}:
Embedding Average (\textbf{Average}), Embedding Extrema (\textbf{Extrema}) and Embedding Greedy (\textbf{Greedy}).
To measure informativeness and diversity of the response, we exploited the Distinct-1, Distinct-2 and Distinct-3 metrics.
A higher ratio of distinct n-grams implies more informative and diverse responses.

\subsection{Implementation and Reproducibility}
We implemented our model with ParlAI~\citep{miller2017parlai}.
The sequence lengths are truncated at 50.
We used Adam~\citep{DBLP:journals/corr/KingmaB14} with an initial learning rate of 0.001 to optimize the model.
For all the experiments, we employed a 2-layer bidirectional LSTM as the encoder and a unidirectional one as the decoder.
The hidden size and the word embedding dimension are both set to 300.
The latent variable size is set to 64. 
The topic number $K$ in our model is set to 5 and 
the most frequent 3,159 words are taken as the topical words vocabulary by stemming, filtering stop-words from the training set.
The batch size is set to 128 for all models.
We trained a Twitter LDA model to obtain the topical words for TA-SEQ2SEQ and set its model-specific parameters following the original paper~\cite{DBLP:conf/aaai/XingWWLHZM17}.
For regularization and preventing over-fitting, a dropout of 0.1 is applied and the weight decay is set to $3\times{}10^{-5}$.
We used the pretrained word embeddings~\cite{pennington2014glove} of 300 dimensions,
and the vocabulary size is set to 20,000.
All models are trained with early stopping, i.e., if the loss does not decrease after 10 validations.
The loss is computed on the validation set at every 0.5 epochs and we save the parameters for the top model on the validation set.
We finally report evaluation scores on the test set from the saved model.

\subsection{Overall Performance}
Table~\ref{tbl:main_exp_res} lists the performance of our system and the comparison systems.
CVAE and LAED inject SEQ2SEQ with stochastic latent variable, resulting in more informative responses and better performance on Distinct-\{1, 2, 3\}.
TA-SEQ2SEQ incorporates SEQ2SEQ with the outsourcing topic information from LDA. It is not surprising that it performs much better on the response relevance (BLEU, Average, Greedy, Extrema), while its improvements on the informativeness are limited.
DOM-SEQ2SEQ builds multiple domain-specific encoder-decoders. It gains improvements on both the relevance metrics and informativeness metrics. 

In general, with both the context-aware and topic-aware parameterization, our model outperforms all the competitive baselines in terms of the response relevance and informativeness. 

\subsection{Context-aware \textit{vs} Topic-aware Parameterization}
Context-aware parameterization captures local semantics of the given context, while topic-aware parameterization enables parameters sharing   among conversations with similar topics. 
As shown in Table~\ref{tbl:main_exp_res}, both parameterization mechanisms perform much better than the original SEQ2SEQ model, while context-aware parameterization is slightly better in terms of informativeness. 
When jointly utilizing both the context-aware and topic-aware parameterization mechanisms, we observe the best performance, indicating that these two mechanisms are both beneficial and complementary. 

\begin{table}[!t]
\small
\centering
\scalebox{0.8}{
\begin{tabular}{@{}lcccc@{}}
\toprule
\textbf{Opponent}   & \textbf{Win} & \textbf{Loss} & \textbf{Tie} & \textbf{Kappa}  \\ \midrule
\textbf{\textsc{AdaND} vs. SEQ2SEQ}      & 62.5\%  & 2.52\% & 34.98\% & 46.46  \\  
\textbf{\textsc{AdaND} vs. CVAE}         & 36.98\% & 7.56\% & 55.46\% & 51.59  \\  
\textbf{\textsc{AdaND} vs. LAED}         & 44.54\% & 6.72\% & 48.74\% & 55.57  \\  
\textbf{\textsc{AdaND} vs. TA-SEQ2SEQ}   & 54.62\% & 5.88\% & 39.5\%  & 54.91  \\  
\textbf{\textsc{AdaND} vs. DOM-SEQ2SEQ}  & 32.78\% & 10.08\%& 57.14\% & 55.53  \\ \bottomrule 
\end{tabular}
}
\caption{The results of human evaluation.}
\label{tbl:human_eval}
\end{table}

\begin{table}[!t]
\centering
\small
\scalebox{0.90}{
\begin{tabular}{lc}
\toprule
\textbf{Models}      & \textbf{Speed (cases/ms)} \\
\midrule
\textbf{SEQ2SEQ}     &      0.740   \\
\textbf{CVAE}        &      0.541   \\
\textbf{LAED}        &      0.435   \\
\textbf{TA-SEQ2SEQ}  &      0.172   \\
\textbf{DOM-SEQ2SEQ} &      0.106   \\
\textbf{\textsc{AdaND}}       &      0.407   \\
\bottomrule
\end{tabular}
}
\caption{Speed test.}
\label{tbl:speed}
\end{table}

\subsection{Human Evaluation}

We conducted human evaluations on the test set to further validate the effectiveness of the model.
We randomly selected 500 samples from the test set. 
Three well-educated students were invited to conduct the evaluation.
For each case, we provided annotators with triplets (sample, $\text{response}_1$, $\text{response}_2$) whereby one response is generated by \textsc{AdaND}, and the other is generated by a competitor model.
The annotators, who have no knowledge about which system the response is from, are then required to independently rate among win ($\text{response}_1$ is better), loss ($\text{response}_2$ is better) and tie (they are equally good or bad),
considering four factors: context relevance, logical consistency, fluency and informativeness.
Note that if annotators rate different options, this triplet will be counted as ``tie''.
Table~\ref{tbl:human_eval} reveals the results of subjective evaluation.
The kappa scores indicate that the annotators came to a fair agreement in the judgment.

As expected, \textsc{AdaND} outperforms the other baselines and enjoys a large margin over the existing models.
The relative performance of the competitors is consistent with the quantitative evaluation results, confirming the superior performance of our proposed method. 

\subsection{Speed Test}
 We conducted speed test to verify the efficiency of the \textsc{AdaND} model empirically in Table \ref{tbl:speed}.
 Augmented with auxiliary components, all the extension models exhibit higher time cost than the original SEQ2SEQ model.
 \changed{We observe that the decoding speeds of CVAE and LAED are relatively comparable with our model.
 However, when comparing with TA-SEQ2SEQ and DOM-SEQ2SEQ that also elaborately and explicitly model conversations with diverse topics or themes}, \textsc{AdaND} shows a clear superiority in decoding speed.
 For TA-SEQ2SEQ, it relies on an outside LDA model to obtain the topic information. The joint attention and copying mechanism also reduce its efficiency.
 For DOM-SEQ2SEQ, it is not surprising that the time complexity of multiple topic/theme-specific encoder-decoders is much higher than all-other comparison models.  
 \textsc{AdaND} utilizes a single encoder-decoder and is parameterized dynamically regarding the input context, which ensures its flexibility and efficiency.

\begin{table}[!t]
\centering
\scalebox{0.84}{
\begin{tabular}{ccccc}

\toprule
 \textbf{Movie} & \textbf{Politics}  & \textbf{Ubuntu} & \textbf{Food} & \textbf{Network} \\ \midrule
 imax & trump  & install & food  & router \\
 movie & people & grub & vegetarian  & wireless \\ 
 youtube & hillary  & apt & seafood & ip \\
 scene & vote  & kernel & restaurants & address\\
 marvel & clinton  & nvidia & cotto & phone \\
 hulk & election  & cd & gourmet & network \\
 avengers & debate  & sudo & serves & card \\
 nolan & donald  & ssh & starbucks & eth0 \\
 comics & support & boot & breakfast & dhcp \\
 batman & working  & ubuntu & pizzeria & wifi \\
\bottomrule

\end{tabular}
}
\caption{Topics by the words~($\beta$ in Eq.(\ref{eq:beta_theta})) with top-10 highest probability discovered by the latent topic inferrer.}
\label{tbl:topics_by_words}
\vspace{-0.5cm}
\end{table}
\begin{figure}[!t]
  \centering
  \includegraphics[width=0.4\textwidth]{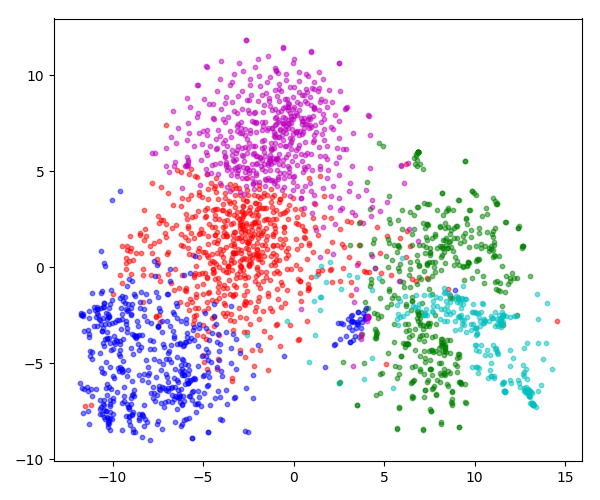}
  \caption{
    t-SNE~\citep{vanDerMaaten2008} projection of topical word embeddings~($\digamma$ in Eq.(\ref{eq:beta_k})) in the latent topic inferrer.
    Words with similar topics are in the same color.
  }
  \label{fig:tsne}
\end{figure}

\definecolor{mycolor}{HTML}{2B83BA}

\begin{table}[!hbt]
\centering
\small
\scalebox{0.84}{
\begin{tabular}[t]{p{0.02cm}|p{8.2cm}}
\toprule

\multirow{9}{*}{1}
&\textbf{\textit{Context:}}   \textit{Hi, can you help me find a good movie to watch?} \\
&\textbf{\textit{Reference:}}   \textit{Hi, I would like to help out.}\\
&\textbf{\textit{SEQ2SEQ:}}    \textit{I'm sorry, I'm sorry, I'm sorry.}\\
&\textbf{\textit{CVAE:}}   \textit{I'm looking for a good comedy movie.}\\
&\textbf{\textit{LAED:}}   \textit{I'm sorry, I'm not sure.}\\
&\textbf{\textit{TA-SEQ2SEQ:}}    \textit{I am looking for a good movie.}\\
&\textbf{\textit{DOM-SEQ2SEQ:}}   \textit{I sent the info on your screen, drive carefully!}\\
&\textbf{\textsc{AdaND}:}   \textit{What kind of movies do you like?}
\\ 
&\textbf{\textit{Inferred topic distribution:}}
\begin{tikzpicture}[thick,scale=0.8, every node/.style={scale=1}]
\begin{axis}[
    ybar,
    symbolic x coords={topic1,topic2,topic3,topic4,topic5},
    ticks=none,
    axis line style={draw=none},
    tick style={draw=none},
    yticklabels={,,},
    width=6.0cm,height=2.2cm,
    bar width=14,
    enlarge x limits=0.3,
    ]
\addplot[mycolor, fill=mycolor] coordinates 
{(topic1,0.501) (topic2,0.02108) (topic3,0.0878) (topic4,0.145) (topic5,0.245)};
\end{axis}
\end{tikzpicture}
\\
\midrule

\multirow{9}{*}{2}
&\textbf{\textit{Context:}} \textit{I love to sit on the couch and play video games myself.} \\
&\textbf{\textit{Reference:}}   \textit{My brother does too, especially since our parents moved to florida recently.}\\
&\textbf{\textit{SEQ2SEQ:}}    \textit{I'm sorry, I'm not a good.}\\
&\textbf{\textit{CVAE:}}  \textit{I'm not sure that is my favorite.}\\
&\textbf{\textit{LAED:}}   \textit{I'm not a good. I'm a good.}\\
&\textbf{\textit{TA-SEQ2SEQ:}}    \textit{That's cool.}\\
&\textbf{\textit{DOM-SEQ2SEQ:}}   \textit{I'm a good. I'm a \_\_unk\_\_.}\\
&\textbf{\textsc{AdaND}:}   \textit{That's cool. What do you do for a living?}
\\  
&\textbf{\textit{Inferred topic distribution:}}
\begin{tikzpicture}[thick,scale=0.8, every node/.style={scale=1}]
\begin{axis}[
    ybar,
    symbolic x coords={topic1,topic2,topic3,topic4,topic5},
    ticks=none,
    axis line style={draw=none},
    tick style={draw=none},
    yticklabels={,,},
    width=6.0cm,height=2.2cm,
    bar width=14,
    enlarge x limits=0.3,
    ]
\addplot[mycolor, fill=mycolor] coordinates 
{(topic1,0.301) (topic2,0.22108) (topic3,0.0878) (topic4,0.145) (topic5,0.245)};
\end{axis}
\end{tikzpicture}
\\
\midrule

\multirow{9}{*}{3}
&\textbf{\textit{Context:}} \textit{We go back to celebration in Florida often. It is our home town.} \\
&\textbf{\textit{Reference:}}   \textit{Oh how nice! I love taking my dog to the beach!}\\
&\textbf{\textit{SEQ2SEQ:}}    \textit{I'm sorry, I'm not a good.}\\
&\textbf{\textit{CVAE:}}   \textit{I'm a little more than a lot of time.}\\
&\textbf{\textit{LAED:}}   \textit{I'm sorry, I'm sorry, but there are no results for any other restaurants in the centre. Would you like to try another area?}\\
&\textbf{\textit{TA-SEQ2SEQ:}}    \textit{:)}\\
&\textbf{\textit{DOM-SEQ2SEQ:}}   \textit{I have a 4 star hotel near a park, museum, and theatre, 8.86/10 guest rating, free parking, breakfast, and wifi, from September 2nd to 13th, business class flights, \_\_unk\_\_. 86.}\\
&\textbf{\textsc{AdaND}:}   \textit{That's cool. Do you have any hobbies?}
\\ 
&\textbf{\textit{Inferred topic distribution:}}
\begin{tikzpicture}[thick,scale=0.8, every node/.style={scale=1}]
\begin{axis}[
    ybar,
    symbolic x coords={topic1,topic2,topic3,topic4,topic5},
    ticks=none,
    axis line style={draw=none},
    tick style={draw=none},
    yticklabels={,,},
    width=6.0cm,height=2.2cm,
    bar width=14,
    enlarge x limits=0.3,
    ]
\addplot[mycolor, fill=mycolor] coordinates 
{(topic1,0.02652) (topic2,0.1551) (topic3,0.07843) (topic4,0.2841) (topic5,0.4559)};
\end{axis}
\end{tikzpicture}
\\
\midrule

\multirow{9}{*}{4}
&\textbf{\textit{Context:}}  \textit{Oh, that's all right. No need to reserve my ticket. But I do need the travel time, please.} \\
&\textbf{\textit{Reference:}}   \textit{Your travel time is 60 minutes.  Is there anything else that I can do for you? }\\
&\textbf{\textit{SEQ2SEQ:}}   \textit{I have booked you for the reference number is \_\_unk\_\_.} \\ 
&\textbf{\textit{CVAE:}}   \textit{The travel time is 105 minutes.} \\
&\textbf{\textit{LAED:}}  \textit{I' m sorry, I' m sorry, I' m sorry, but there are no results for that time. Would you like to try another day or time?} \\
&\textbf{\textit{TA-SEQ2SEQ:}}  \textit{The train id is \_\_unk\_\_.} \\
&\textbf{\textit{DOM-SEQ2SEQ:}}  \textit{You're welcome. Have a great day!} \\
&\textbf{\textsc{AdaND}:}  \textit{The travel time is 163 minutes. Is there anything else i can help you with?} \\
&\textbf{\textit{Inferred topic distribution:}}
\begin{tikzpicture}[thick,scale=0.8, every node/.style={scale=1}]
\begin{axis}[
    ybar,
    symbolic x coords={topic1,topic2,topic3,topic4,topic5},
    ticks=none,
    axis line style={draw=none},
    tick style={draw=none},
    yticklabels={,,},
    width=6.0cm,height=2.2cm,
    bar width=14,
    enlarge x limits=0.3,
    ]
\addplot[mycolor, fill=mycolor] coordinates 
{(topic1,0.101) (topic2,0.02108) (topic3,0.4878) (topic4,0.145) (topic5,0.245)};
\end{axis}
\end{tikzpicture}
\\

\midrule

\multirow{9}{*}{5}
&\textbf{\textit{Context:}}  \textit{How is gonna be the weather forecast in San Mateo for the next weekend?} \\
&\textbf{\textit{Reference:}}   \textit{In San Mateo it will be stormy on Saturday, and hail on Sunday.}\\
&\textbf{\textit{SEQ2SEQ:}}   \textit{The address is a park, and the address is a park.} \\ 
&\textbf{\textit{CVAE:}}   \textit{It will be a low of 60F and a high of 70F.} \\
&\textbf{\textit{LAED:}}  \textit{what city would you like to know the weather for?} \\
&\textbf{\textit{TA-SEQ2SEQ:}}  \textit{Do you have a budget?} \\
&\textbf{\textit{DOM-SEQ2SEQ:}}  \textit{It will not be cloudy this week.} \\
&\textbf{\textsc{AdaND}:}  \textit{It will not be cloudy in San Mateo next week.} \\
&\textbf{\textit{Inferred topic distribution:}}
\begin{tikzpicture}[thick,scale=0.8, every node/.style={scale=1}]
\begin{axis}[
    ybar,
    symbolic x coords={topic1,topic2,topic3,topic4,topic5},
    ticks=none,
    axis line style={draw=none},
    tick style={draw=none},
    yticklabels={,,},
    width=6.0cm,height=2.2cm,
    bar width=14,
    enlarge x limits=0.3,
    ]
\addplot[mycolor, fill=mycolor] coordinates 
{(topic1,0.2311) (topic2,0.04226) (topic3,0.04115) (topic4,0.137) (topic5,0.5484)};
\end{axis}
\end{tikzpicture}
\\

\bottomrule
\end{tabular}
}
\caption{
Test samples of our model (\textsc{AdaND}) and the baselines.
The latent topic distributions inferred by \textsc{AdaND} are also presented.
The reference is the ground-truth response in the dataset.
}
\label{tbl:case_show}
\end{table}

\subsection{Analysis \& Case Study}

 To get some insights of how topic-aware parameterization performs,
 we present the topics by the words ($\beta$ in Eq.(\ref{eq:beta_theta})) with top-10 highest probabilities in Table~\ref{tbl:topics_by_words}.
 The discernible clusters of the topical words ($\digamma$ in Eq.(\ref{eq:beta_k})) are illustrated in Figure~\ref{fig:tsne}.
 These evidences demonstrate that the topic inferrer in topic-aware parameterization effectively distills the latent topic distribution of each conversation, which enables the parameter sharing among conversations with similar topics.
 
 \changed{We also investigate the orthogonality of the learned $U$ and $V$ matrices in Eq.(\ref{eq:context_svd}) and Eq.(\ref{eq:topic_svd}).
 We trained our model multiple times with different parameter initialization methods (drawn values from normal distribution or uniform distribution). 
 We observe that $UU^\mathrm{T}$ and $VV^\mathrm{T}$ approximate identity matrices.
 We conjecture that such SVD-alike parameterization implicitly enforces orthogonality during training.
 }
 
 We list several examples generated by different models in Table~\ref{tbl:case_show}.
 The inferred latent topic distributions are also presented in the table.
 It can be observed that responses generated by the original SEQ2SEQ model are more generic.
 Latent variable conversation models (CVAE and LAED) generate more diverse but sometimes irrelevant responses,
 TA-SEQ2SEQ tends to produce short responses while DOM-SEQ2SEQ does not perform obviously better than TA-SEQ2SEQ.
 The responses generated by \textsc{AdaND} are not only relevant but also informative.
\section{Related Work}

 Our work is closely related to the research of dialogue generation in diverse conversations.
 Previous work relies on external pre-organized topic information~\citep{DBLP:conf/aaai/XingWWLHZM17, DBLP:conf/emnlp/WangJBN17} or predicted keywords~\citep{DBLP:conf/emnlp/YaoZFZY17, DBLP:conf/sigir/WangHXSN18} to boost the response informativeness and coherence.
 \citet{DBLP:journals/corr/abs-1708-00897} further leveraged the topic/theme annotations to build multiple separate encoder-decoder models for topic/theme-aware response generation.
 In contrast, we do not exploit any outsourcing or labeled topic information. The proposed model directly infers the latent topics of each conversation and is trained in an end-to-end manner.
 Another difference is that we maintain a single encoder-decoder for various conversations whereas the model is dynamically and specially parameterized.

 The second line of related work is parameterization in NLP.
 \citet{DBLP:journals/corr/HaDL16} proposed to train a small network to generate the parameters for another larger network. 
 Such adaptive parameterization has been shown to be successful in many NLP tasks, including language modeling~\citep{Suarez2017CharacterLevelLM,DBLP:conf/nips/FlennerhagYKE18}, sequence generation~\citep{DBLP:journals/corr/HaE17, peng2019text}, and neural machine translation~\cite{DBLP:conf/emnlp/PlataniosSNM18}.
 In our work, we parameterize the encoder-decoder with respect to both the context and the latent topics.

 Regarding latent variable conversation models, prior researches strive to learn meaningful latent variables for dialogue systems, and reveal that latent variables befit the neural dialogue models with more diverse response generations~\citep{DBLP:journals/corr/SerbanSLCPCB16, DBLP:conf/acl/ZhaoZE17, DBLP:conf/eacl/ClarkC17, chen2018hierarchical} and interpretable dialogue actions~\cite{LAED}.
 In our model, instead of directly injecting the latent variable into dialogue models, we distill the latent topics through neural variational inference, offering a more interpretable latent variable.
 Moreover, we parameterize the encoder-decoder with the inferred latent topics, which allows parameter sharing among conversations with similar topics.
\section{Conclusion}

 This paper presents an adaptive neural dialogue generation model---\textsc{AdaND}, which allows the dynamical parameterization of the model to each conversation and enables the generation of appropriate responses in diverse conversations.
 Specially, we propose two adaptive parameterization approaches: context-aware parameterization which captures local semantics of the input context; and topic-aware parameterization which enables parameter sharing by first inferring the latent topics of the given context and then generating the parameters with the inferred latent topics. 
 The proposed approaches are assessed on a large-scale conversational dataset and the results show that our model achieves superior performance and higher efficiency.
 \changed{It should be noted that our approach is not isolated to only LSTMs.
 We would like to explore the effectiveness of the approach regarding other structures in future work.}

\section*{Acknowledgments}

We would like to thank all the reviewers for their insightful and valuable comments and suggestions.
Hongshen Chen and Cheng Zhang are the corresponding authors.
\bibliography{Main}
\bibliographystyle{acl_natbib}

\end{document}